%% file: acl_latex.tex
\pdfoutput=1

\documentclass[11pt]{article}

\usepackage{amsmath}
\usepackage{amssymb}
\usepackage{makecell}
\usepackage{paralist}
\usepackage{enumitem}
\usepackage{epsfig,subfigure, wrapfig}
\usepackage[section]{algorithm}
\usepackage{algorithmic}
\usepackage{multirow}
\usepackage{booktabs}

\newcommand{\name}{CHESS }

\usepackage[preprint]{acl}

\usepackage{times}
\usepackage{latexsym}

\usepackage[T1]{fontenc}

\usepackage[utf8]{inputenc}

\usepackage{microtype}

\usepackage{inconsolata}

\usepackage{graphicx}

\title{\name: Optimizing LLM Inference via Channel-Wise Thresholding and Selective Sparsification}

\author{
  \textbf{Junhui He\textsuperscript{1,2}}, \ 
  \textbf{Shangyu Wu\textsuperscript{3,4}}, \ 
  \textbf{Weidong Wen\textsuperscript{2}}, \ 
  \textbf{Chun Jason Xue\textsuperscript{3}}, \ 
  \textbf{Qingan Li\textsuperscript{2 \Thanks{Corresponding author}}}
  \\
  \textsuperscript{1} Key Laboratory of Aerospace Information Security and Trusted Computing, Ministry \\
  of Education, School of Cyber Science and Engineering, Wuhan University,
  \\
  \textsuperscript{2} School of Computer Science, Wuhan University,
  \\
  \textsuperscript{3} MBZUAI,
  \\
  \textsuperscript{4} City University of Hong Kong
}

\begin{document}
\maketitle

\begin{abstract}
\input{sections/0-abstract}
\end{abstract}



\input{sections/1-intro}
\input{sections/2-background}
\input{sections/3-method}
\input{sections/4-experiments}
\input{sections/6-rel}

\input{sections/5-conclusions}
\input{sections/7-lim}
\input{sections/8-ack}

\bibliography{sections/references}  






\end{document}

%% file: sections/0-abstract.tex
Deploying large language models (LLMs) on edge devices presents significant challenges due to the substantial computational overhead and memory requirements. 
Activation sparsification can mitigate these resource challenges by reducing the number of activated neurons during inference. 
Existing methods typically employ thresholding-based sparsification based on the statistics of activation tensors. 
However, they do not model the impact of activation sparsification on performance, resulting in suboptimal performance degradation.
To address the limitations, this paper reformulates the activation sparsification problem to explicitly capture the relationship between activation sparsity and model performance.
Then, this paper proposes \textbf{\name}, a general activation sparsification approach via \textbf{CH}annel-wise thr\textbf{E}sholding and \textbf{S}elective \textbf{S}parsification.
First, channel-wise thresholding assigns a unique threshold to each activation channel in the feed-forward network (FFN) layers. 
Then, selective sparsification involves applying thresholding-based activation sparsification to specific layers within the attention modules. 
Finally, we detail the implementation of sparse kernels to accelerate LLM inference.
Experimental results demonstrate that the proposed \name achieves lower performance degradation over eight downstream tasks while activating fewer parameters than existing methods, thus speeding up the LLM inference by up to 1.27x.

%% file: sections/1-intro.tex
\section{Introduction}

Large Language Models (LLMs) have become integral to diverse applications, including code generation, office assistance, voice assistance, and assistive applications for individuals with disabilities.
However, due to the substantial computation and memory requirements of LLM inferences, deploying LLMs on edge devices is still challenging.
To mitigate these overheads, utilizing the inherent activation sparsity of LLM has emerged as a promising strategy~\citep{deja_vu, powerinfer, llm_in_a_flash}. 
This approach has proven effective for models with the ReLU activation function~\citep{lazy_neuron, deja_vu}.

Contemporary LLMs demonstrate that SwiGLU or GeGLU activation functions can further boost the model performance, but they induce less activation sparsity.
Consequently, several methods~\citep{relu_strikes_back, prosparse} are proposed to increase sparsity by applying regularization to the activation function and employing continuous training.
However, those works require fine-tuning the LLMs, which entails significant overheads.
To avoid these overheads and increase activation sparsity in modern LLMs, \citet{cats} propose a thresholding-based pruning method to actively sparsify the activation tensors during the inference stage. 
However, this thresholding technique focuses solely on the statistics of the activation tensors themselves, failing to model the impact of sparsification on overall model performance, which results in suboptimal performance degradation.

To address the above limitations, this paper proposes \textbf{\name}, a new activation sparsification optimization via \textbf{CH}annel-wise thr\textbf{E}sholding and \textbf{S}elective \textbf{S}parsification.
First, this paper reformulates the activation sparsification problem to explicitly capture the relationship between activation sparsity and model performance.
Then, this paper proposes channel-wise thresholding for FFN modules in LLMs, which determines a unique threshold for each activation channel.
Furthermore, this paper proposes selective sparsification, which applies thresholding-based activation sparsification to the target layers in the attention module.
Finally, this paper presents the implementations of sparse kernels to accelerate the inference based on the sparse activations.

To validate the effectiveness of the proposed \name, this paper conducts comprehensive experiments on various downstream tasks and state-of-the-art LLMs.
Experimental results demonstrate that \name  can achieve lower performance degradation with better end-to-end inference speedup.
The code is publicly available~\footnote{https://github.com/ZeonfaiHo/CHESS}.

The main contributions of this paper are,
\begin{itemize}[noitemsep, topsep=0pt]
    \item This paper reformulates the activation sparsification problem and establishes a connection between sparsity and model performance.
    \item This paper proposes two activation sparsification methods, the channel-wise thresholding for FFN modules and the selective sparsification for Attention modules, which can be widely applied in existing LLMs.
    \item To make full use of the activation sparsity, this paper details the algorithm design and implementations of the efficient sparse kernels.
    \item Experimental results demonstrate the performance and efficiency of the proposed \name.
\end{itemize}

%% file: sections/2-background.tex
\section{Background and Motivations}

\subsection{Activation Sparsification}

Activation functions introduce non-linearity into neural networks, allowing networks to capture complex patterns in the data.
ReLU~\citep{relu}, as a popular activation function, has been widely applied in most neural networks to address gradient vanish issues~\citep{opt}.
Another benefit of ReLU is introducing the sparsity into the activation tensors.
Recent studies~\citep{lazy_neuron, deja_vu} have demonstrated that up to 95\% of the intermediate FFN activations in OPT models are zero.
Such sparsity can be used to accelerate the model inference while maintaining comparable model performance~\citep{deja_vu, llm_in_a_flash, powerinfer}.

Recent state-of-the-art LLMs replace the ReLU activation function with more advanced activation functions, such as GeLU~\citep{gelu}, SiLU~\citep{silu}, or GLU-series functions~\citep{glu}. 
Although these activation functions can significantly boost the LLMs' performance~\citep{llama}, they induce less activation sparsity.
Previous optimizations based on activation sparsity may not be suitable for the LLMs with those activation functions.

To increase activation sparsity in modern LLMs, existing work~\citep{cats} proposes a thresholding-based pruning method called CATS on some activation tensors in FFN layers.
CATS first computes the cutoff threshold over a set of sample input data according to the given sparsity level, then sparsifies the activations during inference and achieves end-to-end speedup via efficient sparse kernel design.
Although CATS increases activation sparsity, it only focuses on the statistics of the activation tensors without modeling the impact of activation sparsification on the model performance, leading to suboptimal performance drop.

\subsection{Motivation}

\label{sec:motivation}

Following the observations in CATS~\citep{cats}, this paper also aims to apply activation sparsification in the gated-MLP blocks of FFN modules.
The formal expression of the FFN module is defined as,
\begin{equation}
\text{FFN}(x) = \left(\sigma(xW^{\text{gate}}) \odot (xW^{\text{up}})\right) W^{\text{down}}
\end{equation}
where $W^{\text{up}}$, $W^{\text{gate}}$, $W^{\text{down}}$ are parameters, and $\sigma (\cdot)$ is the activation function.
Therefore, the activation values in gated-MLP blocks are,
\begin{equation}
a^{\text{up}}=xW^{\text{up}}, \quad a^{\text{gate}}=\sigma(xW^{\text{gate}})
\end{equation}

Since the activation function introduces sparsity where the values of many elements in the output tensor are close to zero, we focus on pruning the output of the gate projection layer, i.e., $ a^{\text{gate}} $.
Then, the following computations, such as the matrix multiplication for $ a^{\text{up}} $, the element-wise multiplication between $ a^{\text{up}} $ and $ a^{\text{gate}} $, or the matrix multiplication with $ W^{\text{down}} $, can further be optimized due to the zero elements in the pruned $ a^{\text{gate}} $.

Inspired by layer-wise weight pruning ~\citep{sparsegpt, wanda}, this paper reformulates the activation sparsification problem to \textbf{find the optimal pruned activation tensor \( \hat a^{\text{gate}} \) that guarantees a specified sparsity level while minimizing the output difference of the succeeding layer before and after pruning.}
More formally, the problem is defined as,
\begin{equation}
\arg\min_{\hat{a}^{\text{gate}}} \left\|a^{\text{up}}\odot a^{\text{gate}}-a^{\text{up}}\odot \hat a ^{\text{gate}}\right\|^2_2
\label{eq:layerwise_sparsification}
\end{equation}
where $a^{\text{up}}, a^{\text{gate}}$ are different activation tensors in FFN layers, $\hat{a}^{\text{gate}}$ is the pruned activation tensor.

We decompose all activations in the pruned tensor into two subsets, i.e., the pruned $\hat{a}^{\text{gate}}_{\mathcal{P}}$ which are all zeros and the non-pruned $\hat{a}^{\text{gate}}_{\mathbb{U}-\mathcal{P}}$ which keep the original values in $a^{\text{gate}}$.
Thus, we can simplify the objective defined in Equation \ref{eq:layerwise_sparsification} as: \textbf{finding a subset of indices $\mathcal{P}$ that indicates the index of the pruned elements, and satisfies sparsity level $|\mathcal{P}| \ge k \cdot |\mathbb{U}|$, while minimizing the sparsification error illustrated in Equation~\ref{eq:final_obj}}, where $\mathbb{U}=\{1,\ldots, d\}$, $d$ is the feature dimension of $a^{\text{gate}}$.

\begin{equation}
\label{eq:final_obj}
\arg\min_{\mathcal{P}} \sum_{i \in \mathcal{P}} \left(a^{\text{up}}_{i} a^{\text{gate}}_{i}\right)^2
\end{equation}

Equation \ref{eq:final_obj} establishes the theoretical relationship between activation sparsification and model performance, which is ignored by previous works, e.g., CATS \citep{cats}.
However, finding the top-k smallest elments of $\left(a^{\text{up}}_ia^{\text{gate}}_i\right)^2$ requires the prior compuation of $ a^{\text{up}} $, which is not all necessary.
Besides, sorting across channels in each FFN layer is also a costly process.
These challenges point us to propose the CHESS method.

%% file: sections/3-method.tex
\section{\name: Activation Sparsification via Channel-Wise Thresholding and Selective Sparsification}

\begin{figure*}[t]
\centering
\begin{tabular}{cc}
\subfigure[channel 1241] {
    \includegraphics[width=0.3\textwidth]{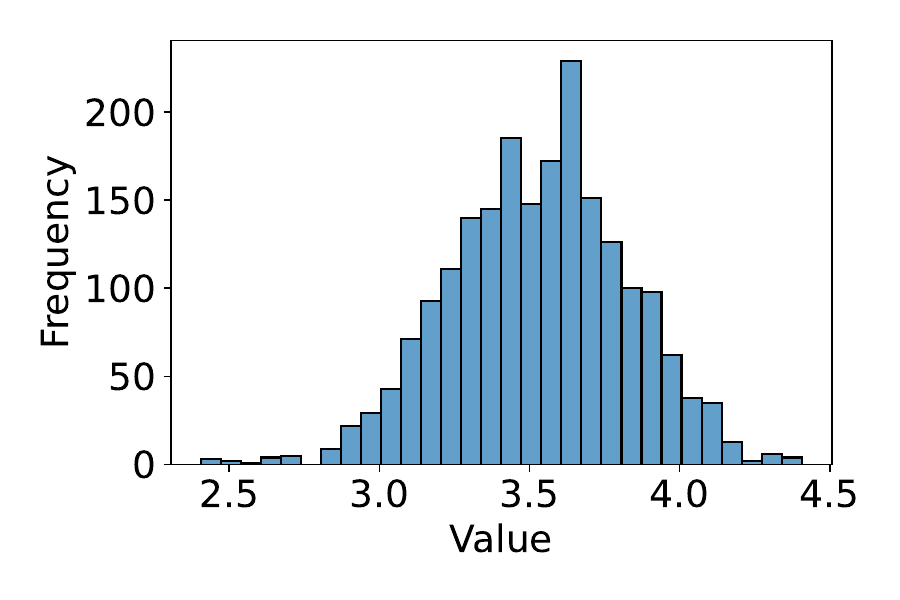}
}
\subfigure[channel 8718]{
    \includegraphics[width=0.3\textwidth]{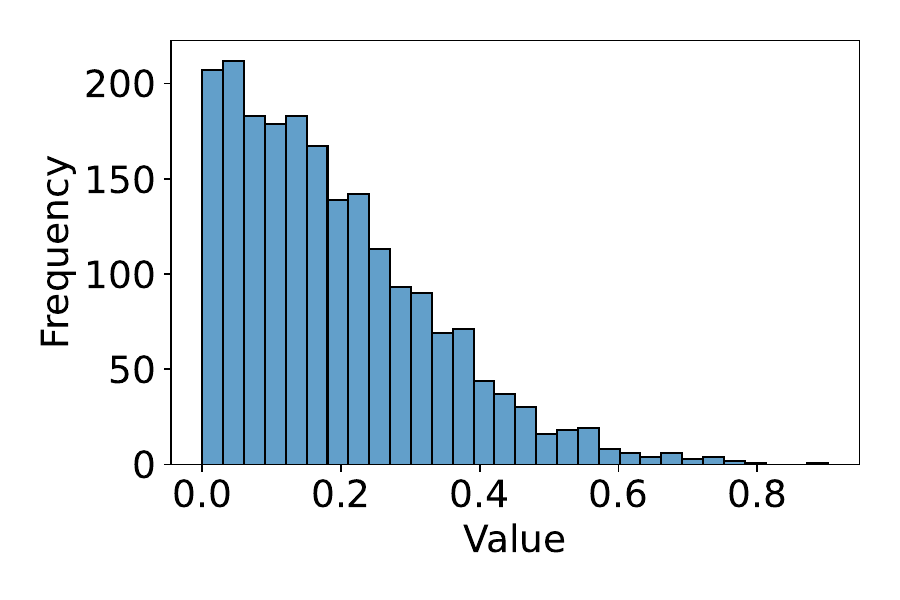}
}
\subfigure[channel 12005]{
    \includegraphics[width=0.3\textwidth]{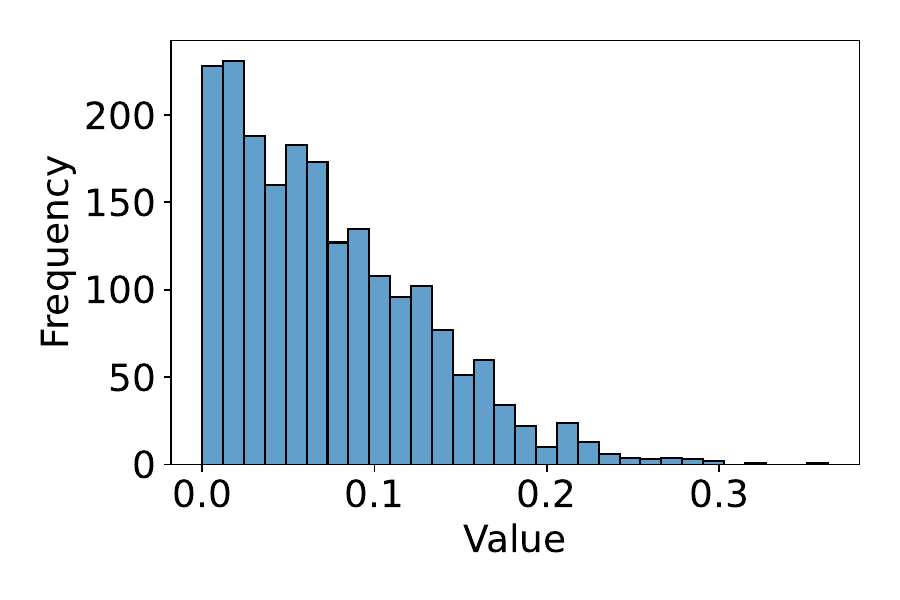}
}
\end{tabular}
\caption{
Distribution of absolute activation values $ | a^{\text{up}}_i | $ across different inputs for various channels in the FFN of layer 16 of the Llama-3-8B model.
}
\label{fig:channel_abs_mean}
\end{figure*}

\begin{figure*}[t]
\centering
\begin{tabular}{cc}
\subfigure[query projection]{
    \includegraphics[width=0.23\textwidth]{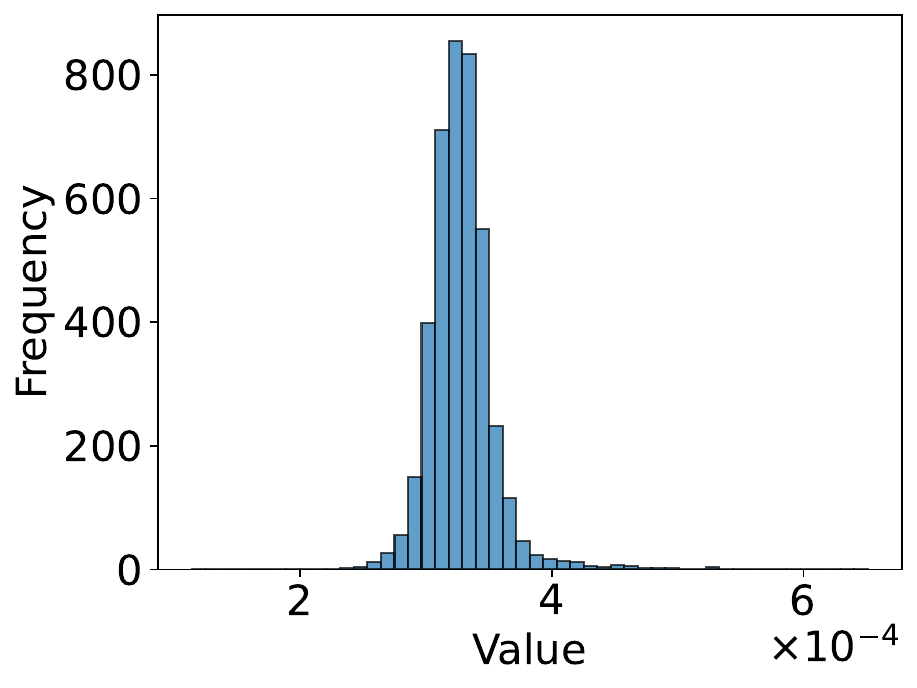}
}
\subfigure[key projection]{
    \includegraphics[width=0.23\textwidth]{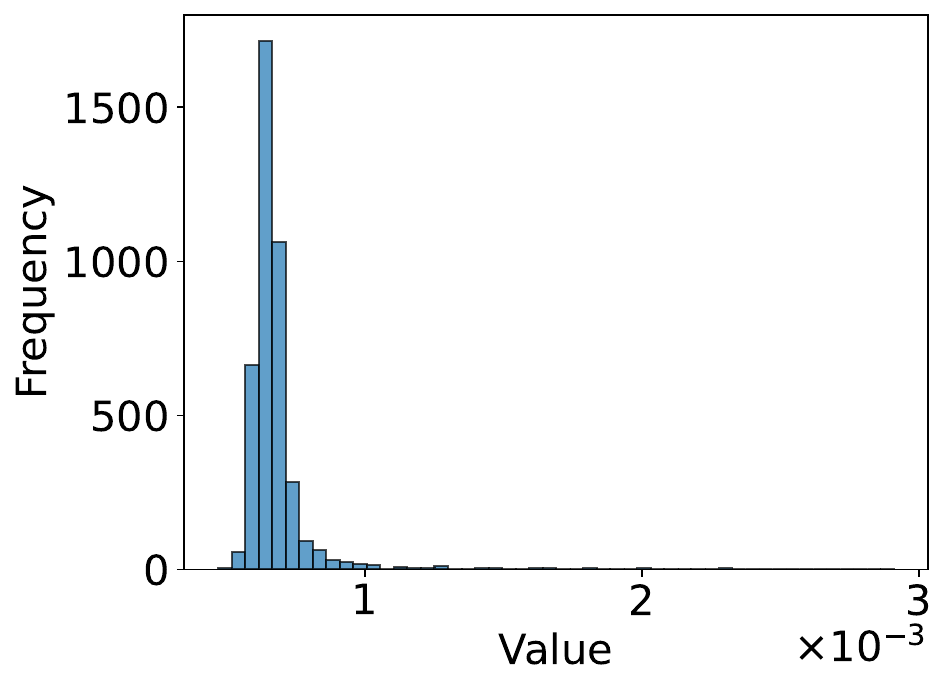}
}
\subfigure[value projection]{
    \includegraphics[width=0.23\textwidth]{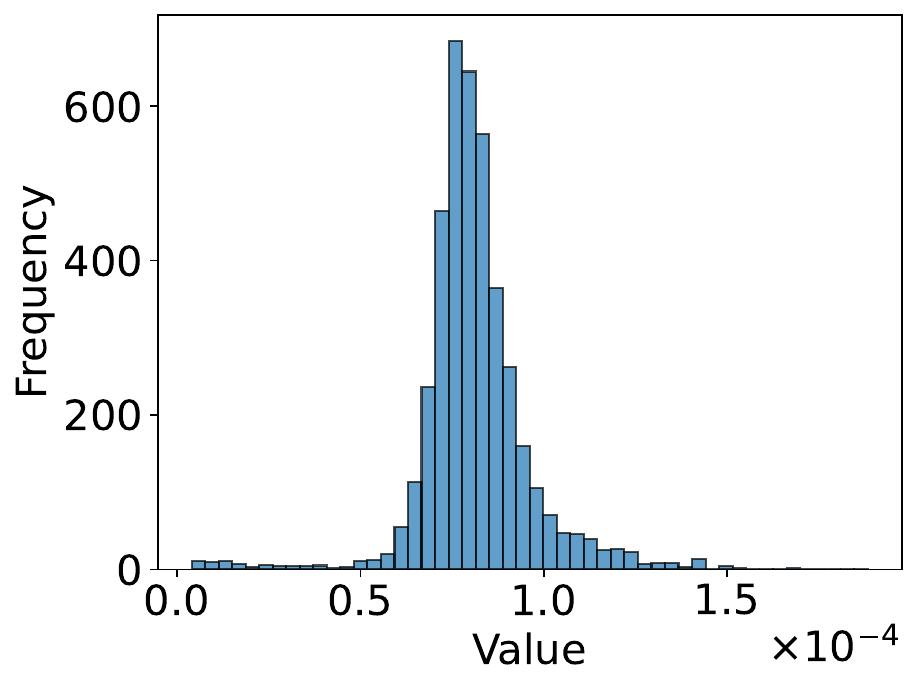}
}
\subfigure[output projection]{
    \includegraphics[width=0.23\textwidth]{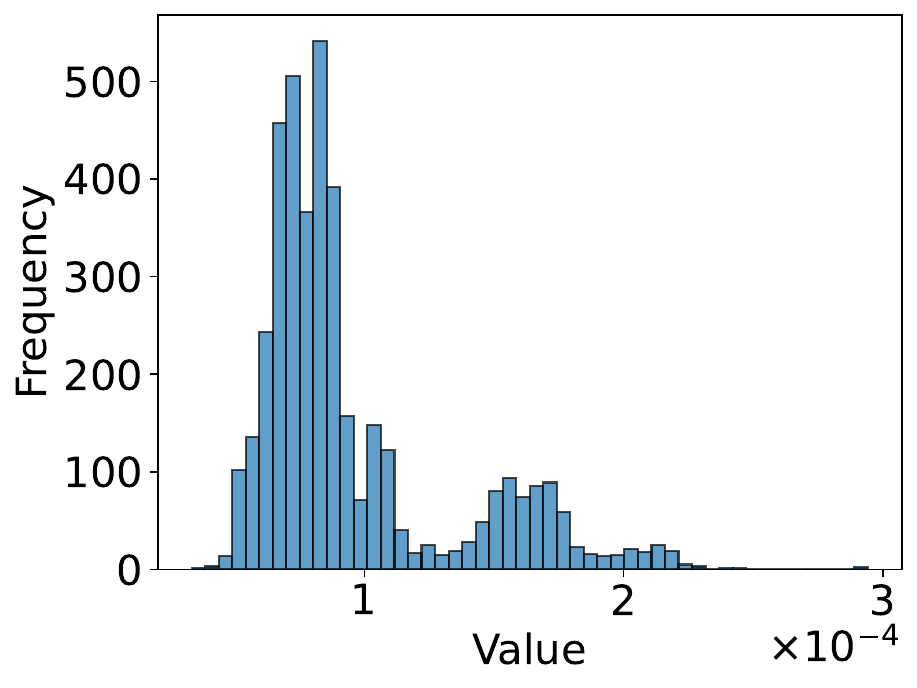}
}
\end{tabular}
\caption{
Distribution of \( \| W_{i,:} \|_2^2 \) of different rows \( i \) in attention projections of layer 16 of Llama-3-8B. 
}
\label{fig:row_norm_distribution}
\end{figure*}

In this section, this paper first introduces the channel-wise thresholding method for FFN modules.
Then, it presents the selective sparsification for attention modules.
Finally, it discusses the algorithm design and implementations of the efficient custom sparse kernels.

\subsection{Channel-Wise Thresholding}
\label{sec:channel_wise_thresholding}

As described in Equation~\ref{eq:final_obj}, whether to prune an activation element is determined by both $a^{\text{up}}$ and $a^{\text{gate}}$.
To quantify the significance of each activation element, we introduce the \textit{importance score} based on Equation \ref{eq:final_obj},
\begin{equation}
\label{eq:score}
\text{score}_i = \left|a_i^{\text{up}}a^{\text{gate}}_i\right|
\end{equation}

To obtain all importance scores of elements in Equation \ref{eq:score}, we need to compute two matrix-multiplication for $ a^{\text{gate}} $ and $ a^{\text{up}} $. 
However, we can reduce the computational cost for $ a^{\text{up}} $ by leveraging the sparsity of $ a^{\text{gate}} $. Therefore, we need to calculate the score only with $ a^{\text{gate}} $.
We observe  that, for each channel $i$, the values of $ |a^{\text{up}}_i| $ remain relatively consistent across different inputs, as shown in Figure \ref{fig:channel_abs_mean}. 
However, these values can vary significantly between different channels.
Based on this observation, this paper estimates the $|a^{\text{up}}_i|$ using the expectation of $|a^{\text{up}}_i|$ over the sampled input data,
\begin{equation}
|a^{\text{up}}_i|\approx \mathbb{E}\left[\left|a^{\text{up}}_i\right|\right]=\frac{1}{n}\sum_{j}|a^{\text{up}}_{ij}|
\end{equation}
where $n$ is the number of sampled input data.
Therefore, the importance score is further estimated as,
\begin{equation}
\hat{\text{score}_i} = \mathbb E\left[\left|a^{\text{up}}_i\right|\right]\left|a_i^{\text{gate}}\right|
\end{equation}

For the sorting overhead, this paper also adopts the CDF-based thresholding method following \citet{cats}.
Specifically, we first outline the cumulative distribution function $F$ of the proposed importance score across all channels,
\begin{equation}
F(t)= \mathbb P (\hat{\text{score}} \le t)
\end{equation}
Then, given a sparsity level $k$, we can obtain the threshold $t_i$ for sparsifying the activation elements on channel $i$,
\begin{equation}
t_i = \frac{\arg\min_{t} F(t)\ge k}{\mathbb{E}\left[\left|a^{\text{up}}_i\right|\right]}
\end{equation}

This threshold indicates the maximal activation magnitude that should be pruned as zero.
Different from CATS, this is a Channel-Wise Thresholding (CWT) technique that relates the model performance with the activation sparsity via introducing the \textit{importance score} in Equation~\ref{eq:score}.

Finally, based on the channel-wise thresholds, the activation values can be sparsified as,
\begin{equation}
\text{CWT}(a_i) = 
\begin{cases}
0, & \text{if } |a_i| \leq t_i \\
a_i, & \text{if } |a_i| > t_i
\end{cases}
\label{eq:cwt}
\end{equation}
and the output of the FFN modules is computed as,
\begin{equation}
\text{FFN}_{\text{CWT}}(x) = \left( \text{CWT}(a^{\text{gate}}) \odot a^{\text{up}} \right) W^{\text{out}}
\end{equation}

\subsection{Selective Sparsification}

Although the activation sparsity in attention modules is much lower than that in FFN modules, applying activation sparsification to these modules can still effectively reduce memory access and computational overhead. 
The standard attention mechanism involves four linear projects: query, key, value, and output projection.
Similar to that in FFN modules, the objective of activation sparsification in the attention module is to \textbf{find the optimal pruned activation tensor that guarantees a specified sparsity level while minimizing the output difference of the succeeding layer before and after pruning.}
More formally, the problem is defined as,
\begin{equation}
\arg\min_{\hat{x}} \|xW-\hat{x}W\|_2^2
\end{equation}
where $ W $ is the weight tensor of the projection layer.

The error E$=\|xW-\hat{x}W\|_2^2$ can be approximated using the Taylor series as follows~\citep{optimal_brain_damage, optimal_brain_surgeon, optimal_brain_compression}:
\begin{equation}
    \text{E} = \mathbf g (\hat x - x)^T + \frac 1 2 (\hat x - x) \mathbf{H} (\hat x - x)^T + O(\| \hat x - x \|^3)
    \label{equ:taylor_series}
\end{equation}
where $\mathbf g$ and $\mathbf H$ denote the first-order and second-order derivatives of the error E with respect to $\hat x$, respectively,
\begin{equation}
\mathbf g = \left. \frac{\partial \text{E}}{\partial \hat x} \right|_{\hat x = x} = 0
\end{equation}
\begin{equation}
\mathbf H = \left. \frac{\partial^2 \text{E}}{\left( \partial \hat x \right)^2} \right|_{\hat x = x} = WW^T
\end{equation}

Then, we replace $\mathbf g$ and $\mathbf H$ with true values, discard the higher-order terms, and apply diagonal approximation to $\mathbf H$. 
The Equation \ref{equ:taylor_series} can be simplified as:
\begin{equation}
    \text{E} \approx \sum_{i = 1}^d \| W_{i,:} \|_2^2 (\hat x_i - x_i)^2
\end{equation}
where $\| W_{i,:} \|_2^2$ denotes the square of $\ell _2 $ norm of row $i$ in weight matrix $W$.
As described in Section~\ref{sec:motivation}, we can also decompose the input features into pruned features (zeros) and non-pruned features (original values) and then transform the objective as follows,
\begin{equation}
    \label{eq:approx_error}
    \arg\min_{\mathcal{P}} \sum_{i\in\mathcal{P}}  \| W_{i,:} \|_2^2 x_i^2
\end{equation}

To further simplify Equation~\ref{eq:approx_error}, this paper analyzes the statistics of the weight matrix in the attention mechanism.
Figure~\ref{fig:row_norm_distribution} shows the distribution of \( \| W_{i,:} \|_2^2 \) of different rows in projection weights.
From the results, all rows from the same weight exhibit similar  \( \| W_{i,:} \|_2^2 \), therefore we can eliminate this coefficient from Equation~\ref{eq:approx_error} and derive the simplified final objective: 
\begin{equation}
\label{eq:final_sparse_attn}
\arg\min_{\mathcal{P}} \sum_{i\in\mathcal{P}} | x_i |
\end{equation}

Based on Equation~\ref{eq:final_sparse_attn}, this paper adopts the tensor-wise thresholiding method proposed by CATS~\citep{cats} for the projection layers of attention modules:
\begin{equation}
\text{CATS}(x_i) = 
\begin{cases}
    0, \quad &\text{if} \ |x_i| \le t \\
    x_i, \quad &\text{if} \ |x_i| > t 
\end{cases}
\label{eq:cats}
\end{equation}

However, which layers the CATS should be applied to becomes a challenge in terms of the trade-off between model performance and model efficiency.
The search space is quite large. 
Taking Llama-2-7B as an example, which has 32 layers and four attention projections per layer, the search space is over the septillion level.

In this paper, we compare two stratagies, namely \textit{full sparsification} and \textit{selective sparsification}.
Full sparsification refers to applying CATS to all four projections of the attention mechanism,
\begin{equation}
\label{equ:full_sparsification}
\begin{split}
\text{C}_{t_{\text{o}}}
(
\text{Attn}
(
\text{C}_{t_{\text{i}}}(x) W^{\text{q}},
\text{C}_{t_{\text{i}}}(x) W^{\text{k}}, 
\text{C}_{t_{\text{i}}}(x) W^{\text{v}}
)
) W^{\text{o}}
\end{split}
\end{equation}
where $C(\cdot)_t$ is the CATS function with the threshold $t$.
Conversely, selective sparsification refers to applying the CATS function to only query and output projections, while not altering key and query projections.
The formal expression is,
\begin{equation}
\label{equ:selective_sparsification}
\text{C}_{t_{\text{o}}}(\text{Attn}(\text{C}_{t_{\text{q}}}(x)W^{\text{q}}, xW^{\text{k}}, xW^{\text{v}}))W^{\text{o}}
\end{equation}

Experimental results (ref. Section 
\ref{sec:fs_vs_ss}) demonstrate that selective sparsification results in significantly lower performance degradation, while achieving comparable overhead reduction when applied to GQA modules. 
Since the GQA modules are widely applied in modern LLMs, we utilize selective sparsification as our main method for attention modules.

\subsection{Efficient Sparse Kernels}

\begin{algorithm}[tbh]
\caption{\textit{spvmm} (sparse vector-matrix multiplication) kernel}
\label{alg:spvmm}
\begin{algorithmic}[1]
\REQUIRE The sparse input vector $x \in \mathbb R^{1 \times K}$, the weight matrix $W \in \mathbb R^{K \times N}$, the number of output elements $N$, the number of input elements $K$, the block size $B$.
\ENSURE The output vector $y \in \mathbb{R}^{1 \times N}$
\FOR{$n0$ from $0$ to $N$ with step size $B$ in \text{PARALLEL}}
    \FOR{$k$ from $0$ to $K$}
        \IF{$x[k] \neq 0.0$}
            \STATE $n1_{upp} = \min(B, N - n0)$
            \FOR{$n1$ from $0$ to $n1_{upp}$ VECTORIZED}
                \STATE $y[n0 + n1] \mathrel{+}= x[k] \times W[k][n0 + n1]$
            \ENDFOR
        \ENDIF
    \ENDFOR
\ENDFOR
\RETURN y
\end{algorithmic}
\end{algorithm}

\begin{algorithm}[tbh]
\caption{\textit{vmmsp} (vector-matrix multiplication with output sparsity) kernel}
\label{alg:vmmsp}
\begin{algorithmic}[1]
\REQUIRE The input vector $x \in \mathbb{R}^{1 \times K}$, the weight matrix $W \in \mathbb{R}^{N \times K}$, the output mask vector $\text{mask} \in \mathbb{R}^{1 \times N}$, the number of output elements $N$, the number of input elements $K$, the block size $B$.
\ENSURE The output vector $y \in \mathbb{R}^{1 \times N}$.
\FOR{$n0$ from $0$ to $N$ with step size $B$ in PARALLEL}
    \STATE $n1_{upp} = \min(B, N - n0)$
    \FOR{$n1$ from $0$ to $n1_{upp}$}
        \IF{$\text{mask}[n0 + n1] \neq 0.0$}
            \STATE $accum = 0.0$
            \FOR{$k$ from $0$ to $K$ VECTORIZED}
                \STATE $accum \mathrel{+}= W[n0 + n1][k] \times x[k]$
            \ENDFOR
            \STATE $\text{y}[n0 + n1] = accum \times \text{mask}[n0 + n1]$
        \ENDIF
    \ENDFOR
\ENDFOR
\RETURN y
\end{algorithmic}
\end{algorithm}

To achieve wall-clock speedup and reduce inference latency via sparse activations, we developed two custom CPU kernels: \textit{spvmm} (sparse vector-matrix multiplication) and \textit{vmmsp} (vector-matrix multiplication with output sparsity). 
The \textit{spvmm} kernel is optimized for cases where the input activation tensor is sparse, and it is employed in attention modules and FFN down projections. 
Conversely, the \textit{vmmsp} kernel is designed for cases where the output activation tensor is multiplied with a sparse mask, and is used in FFN up projections. 

Algorithm~\ref{alg:spvmm} and Algorithm~\ref{alg:vmmsp} show the detailed steps of \textit{spvmm} and \textit{vmmsp}, respectively.
Both algorithms splits the output vector into blocks of size $B$ and accumulates the results of each block in parallel (Line 1 in Algorithm~\ref{alg:spvmm}, Line 1 in Algorithm~\ref{alg:vmmsp}).
What's more, they both reduce the latency by bypassing unnecessary weight reads and computations (Line 3 in Algorithm~\ref{alg:spvmm}, Line 4 in Algorithm~\ref{alg:vmmsp}).
Notably, although the implementation of the \textit{vmmsp} kernel is relatively straightforward, the \textit{spvmm} kernel requires a more complex approach because its access to each column of $W$ is not continuous. 
To address this, we employ two advanced optimizations. 
First, we transpose the linear projection weights in advance during the model preprocessing stage, to ensure memory access continuity.
Additionally, we employ loop tiling and loop reordering to make sure that each threads compute independently without the need for synchronization or atomic operations. 

%% file: sections/4-experiments.tex
\section{Experiments}

In this section, this paper first introduces the dataset, comparisons, and implementation details.
Then, this paper presents the main results over 8 downstream tasks in terms of the model performance and model efficiency.
Besides, this paper also conducts an ablation study across different sparsification methods for the attention module and analysis on performance and efficiency over different sparsity level.
Additionally, this paper conducts extended comparisons with other state-of-the-art training-free pruning methods, to validate the effectiveness of the proposed \name .

\subsection{Datasets and Experimental Setup}

\noindent\textbf{Datasets} 
We utilize ARC Challenge (Arc-C), ARC Easy (Arc-E), BoolQ, HellaSwag (HS), OpenbookQA (QA), PIQA, SCI-Q, Winogrande (WG) as benchmarks for downstream tasks, employing the Evaluation Harness library from Eleuther AI to ensure consistency with \citet{cats}.
These tasks are designed to assess various aspects of the language model's performance, including comprehension, common sense, and reasoning abilities, which effectively illustrate the model's capability loss with activation sparsification.

\noindent\textbf{Comparisons}
To validate the effectiveness of the proposed \name, we conducted experiments using several state-of-the-art LLMs, including Llama-2-7B, Llama-2-13B, Llama-2-70B, Llama-3-8B and Mistral-7B.
These models feature different attention mechanisms, specifically MHA and GQA, and utilize SwiGLU as the activation function for the FFN modules.
We tested four different configurations across all five LLMs:

\begin{itemize}[noitemsep, topsep=0pt]
    \item \textbf{Base Model:} the LLM model without any activation sparsification.
    \item \textbf{CATS~\citep{cats}:} the state-of-the-art activation sparsification method, which applies magnitude pruning to FFN activations.
    \item \textbf{\name w/o:} the proposed method including channel-wise thresholding but without attention sparsification.
    \item \textbf{\name w/:} the channel-wise thresholding and selective sparsification method.
\end{itemize}

For the ablation study, we evaluate the following three models:

\begin{itemize}[noitemsep, topsep=0pt]
    \item \textbf{Llama-3:} the Llama-3 8B model without activation sparsification.
    \item \textbf{FS:} No activation sparsification applied to the FFNs; full sparsification applied in the attention modules.
    \item \textbf{SS:} No activation sparsification applied to the FFNs; selective sparsification applied in the attention modules.
\end{itemize}

\noindent\textbf{Implementation Details}
For all models involving activation sparsification, thresholds are sampled from a subset of the C4 dataset~\citep{c4}. 
Following the settings in CATS~\citep{cats}, the sparsity level $k$ is set to 0.5, where the accuracy drop is minimal while the inference latency significantly decreases. 
The proposed method was implemented using the PyTorch v2.2.2~\citep{pytorch} and HuggingFace Transformers v4.39.3~\citep{transformers}. 
End-to-end decoding speedups are measured on a randomly collected subset of C4 dataset.
Kernel efficiency and end-to-end speedup experiments are conducted with FP32 precision on a personal computer equipped with an Intel Core I9-12900K CPU and 64GB of DDR4 memory.
Since our work can be applied to quantized models as well, changing weight precision to FP16 or even lower bit-width quantizations does not materially affect our results~\citep{cats}.

\begin{table*}[tbh]
\centering
\resizebox{0.92\textwidth}{!}{
\begin{tabular}{lcccccccccc}
\toprule

\textbf{Models} & \textbf{AP$\downarrow$} & \textbf{Arc-C$\uparrow$} & \textbf{Arc-E$\uparrow$} & \textbf{BoolQ$\uparrow$}  & \textbf{HS$\uparrow$}  & \textbf{QA$\uparrow$}  & \textbf{PIQA$\uparrow$}  & \textbf{SciQ$\uparrow$}  & \textbf{WG$\uparrow$}  & \textbf{Avg$\uparrow$}  \\

\midrule

Llama-2-7B  & 100\%     & 43.43 & 76.26 & 77.68 & 57.15 & 31.40 & 78.07 & 93.90 & 69.14 & 65.87 \\
CATS        & 78.16\%   & 41.13 & 74.07 & 72.17 & 57.03 & 31.60 & 77.48 & 92.80 & \textbf{66.69} & 64.12 \\
\name w/o    & 78.17\%  & \textbf{41.47} & \textbf{74.62} & \textbf{74.22} & \textbf{57.15} & 32.40 & 77.20 & 93.20 & 66.61 & \textbf{64.60} \\
\name w/     & \textbf{70.05\%} & 40.36 & 74.37 & \textbf{74.22} & 56.60 & \textbf{33.60} & \textbf{77.86} & \textbf{93.30} & 66.22 & 64.56 \\

\midrule

Llama-2-13B & 100\% & 48.38 & 79.38 & 80.61 & 60.06 & 35.00 & 79.05 & 94.60 & 72.22 & 68.66 \\
CATS & 77.97\% & \textbf{46.93} & 77.44 & 75.60 & 60.42 & 33.80 & 78.78 & 94.10 & 70.64 & 67.21 \\
\name w/o & 77.98\% & 46.67 & \textbf{77.95} & \textbf{79.11} & \textbf{60.64} & 34.00 & 78.89 & \textbf{94.30} & 70.09 & 67.71 \\
\name w/ & \textbf{69.82}\% & 46.84 & \textbf{77.95} & 78.50 & 60.47 & \textbf{34.40} & \textbf{79.00} & 94.20 & \textbf{70.88} & \textbf{67.78} \\

\midrule

Llama-2-70B & 100\% & 54.44 & 82.70 & 83.76 & 64.77 & 37.40 & 82.21 & 96.90 & 77.98 & 72.52 \\
CATS & 72.96\% & \textbf{54.61} & 81.48 & 79.72 & 64.30 & \textbf{37.20} & 81.61 & \textbf{96.10} & \textbf{76.32} & 71.41 \\
\name w/o & 72.97\% & 54.10 & \textbf{81.78} & \textbf{82.17} & \textbf{64.92} & 36.60 & 81.12 & 96.00 & \textbf{76.32} & \textbf{71.63} \\
\name w/ & \textbf{65.24\%} & 54.35 & 81.69 & 81.65 & 64.45 & 36.80 & \textbf{81.77} & \textbf{96.10} & 76.24 & \textbf{71.63} \\

\midrule

Llama-3-8B  & 100\%      & 50.17 & 80.22 & 81.07 & 60.15 & 34.60 & 79.60 & 96.30 & 73.32 & {69.42}          \\
CATS & 74.96\%  & 45.22 & 75.76 & 78.65 & 57.34 & 32.40 & 78.40 & \textbf{94.90} & 70.88 & {66.69} \\
\name w/o & 74.96\%  & \textbf{47.44} & \textbf{77.02} & \textbf{79.97} & \textbf{59.06} & \textbf{32.80} & 78.67 & 94.60 & \textbf{71.90} & \textbf{67.68} \\
\name w/ & \textbf{67.80\%} & 46.67 & 76.85 & 78.04 & 58.62 & \textbf{32.80} & \textbf{79.22} & 94.20 & 70.17 & 67.07 \\

\midrule

Mistral-7B & 100\% & 48.89 & 79.71 & 82.11 & 60.87 & 33.40 & 80.20 & 95.80 & 73.64 & 69.33 \\
CATS & 73.59\% & 48.29 & 77.40 & 79.42 & 60.65 & 31.60 & \textbf{80.52} & 94.40 & 70.48 & 67.85 \\
\name w/o & 73.59\% & 48.21 & \textbf{79.71} & \textbf{80.55} & \textbf{61.70} & 33.20 & 80.41 & \textbf{95.80} & \textbf{70.88} & 68.81 \\
\name w/ & \textbf{66.04\%} & \textbf{49.32} & 79.59 & 80.12 & 61.60 & \textbf{34.40} & 80.20 & 95.00 & 70.56 & \textbf{68.86} \\

\bottomrule
\end{tabular}
}
\caption{Comparison of inference accuracy on downstream tasks of different models. `AP' refers to the ratio of activated parameters.}
\label{tab:performance_comparison}
\end{table*}

\begin{figure}[bt]
\centering
\includegraphics[width=0.9\linewidth]{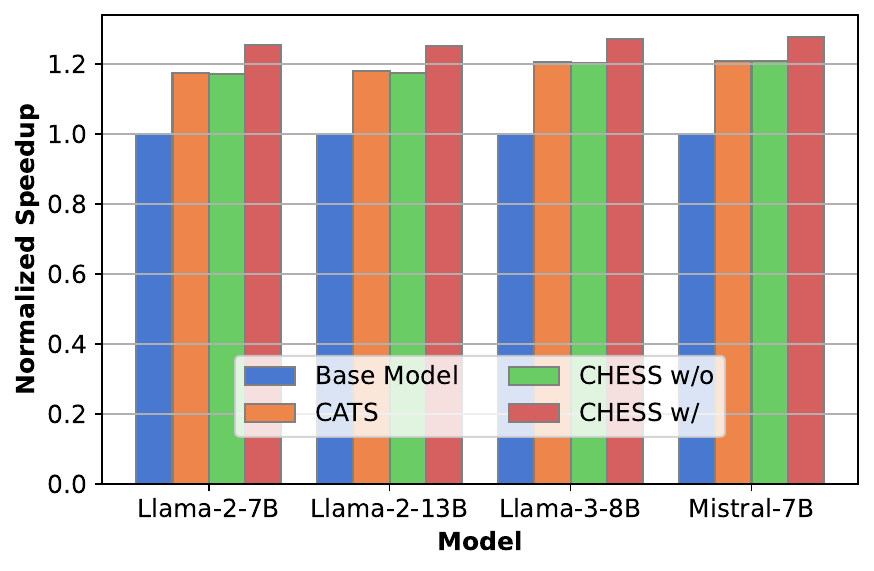}
\caption{End-to-end inference speedup}
\label{fig:end_to_end}
\end{figure}

\begin{table*}[tbh]
\centering
\resizebox{0.92\textwidth}{!}{
\begin{tabular}{lcccccccccc}
\toprule
\textbf{Model} & \textbf{AP$\downarrow$} & \textbf{Arc-C$\uparrow$}  & \textbf{Arc-E$\uparrow$}  & \textbf{BoolQ$\uparrow$}  & \textbf{HS$\uparrow$}  & \textbf{QA$\uparrow$}  & \textbf{PIQA$\uparrow$}  & \textbf{SciQ$\uparrow$}  & \textbf{WG$\uparrow$}  & \textbf{Avg$\uparrow$}  \\
\midrule
Llama-3-8B  & 100\%    & 50.17 & 80.22 & 81.07 & 60.15 & 34.60 & 79.60 & 96.30 & 73.32 & {69.42}          \\
FS & \textbf{90.94\%} & 46.16 & 79.00 & 78.56 & 57.14 & 34.80 & 78.02 & 96.10 & 71.59 & 67.67 \\
SS & 92.84\% & \textbf{50.17} & \textbf{79.67} & \textbf{79.57} & \textbf{59.31} & \textbf{35.00} & \textbf{79.71} & \textbf{96.30} & \textbf{72.85} & \textbf{69.07} \\
\bottomrule
\end{tabular}
}
\caption{Ablation study between full sparsification and selective sparsification in attention modules. `AP' refers to the ratio of activated parameters.}
\label{tab:ablation}
\end{table*}

\begin{table*}[tbh]
    \centering
    \resizebox{0.92\textwidth}{!}{
    \begin{tabular}{lcccccccccc}
        \toprule
        
        \textbf{Model} & \textbf{AP$\downarrow$} & \textbf{Arc-C$\uparrow$}  & \textbf{Arc-E$\uparrow$}  & \textbf{BoolQ$\uparrow$}  & \textbf{HS$\uparrow$}  & \textbf{QA$\uparrow$}  & \textbf{PIQA$\uparrow$}  & \textbf{SciQ$\uparrow$}  & \textbf{WG$\uparrow$}  & \textbf{Avg$\uparrow$}  \\
        
        \midrule

         Llama-3-8B  & 100\%    & 50.17 & 80.22 & 81.07 & 60.15 & 34.60 & 79.60 & 96.30 & 73.32 & {69.42}          \\
         Relufication & 67.10\% & 20.73 & 24.66 & 38.04 & 25.39 & 17.80 & 53.59 & 1.70 & 49.64 & 28.94 \\
         Wanda & \textbf{53.49\%} & 30.80 & 62.58 & \textbf{68.01} & 41.23 & 24.40 & 70.73 & \textbf{91.20} & 62.35 & 56.41 \\
         \name & 54.92\% & \textbf{36.86} & \textbf{67.51} & 66.91 & \textbf{52.92} & \textbf{28.80} & \textbf{75.35} & 89.60 & \textbf{63.69} & \textbf{60.21} \\

         \bottomrule
    \end{tabular}
    }
    \caption{Extended comparisons with state-of-the-art training-free pruning methods. `AP' refers to the ratio of activated parameters; `CHESS' refers to the proposed CHESS model with a sparsity level of 0.7.}
    \label{tab:more_comparison}
\end{table*}

\begin{figure*}[tbh]
\centering
\begin{tabular}{cc}
\subfigure[Attention projection (\textit{spvmm} kernel)]{\includegraphics[width=0.32\linewidth]{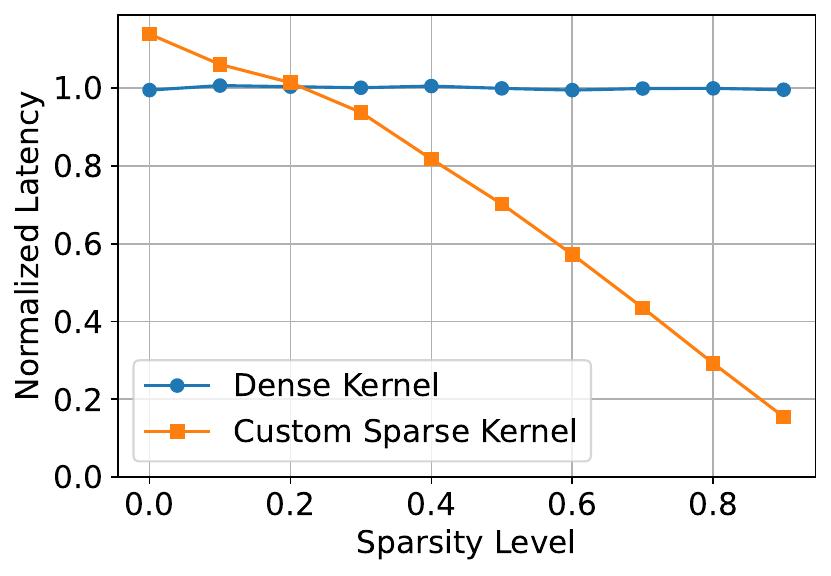}}
\subfigure[Down projection (\textit{spvmm} kernel)]{\includegraphics[width=0.32\linewidth]{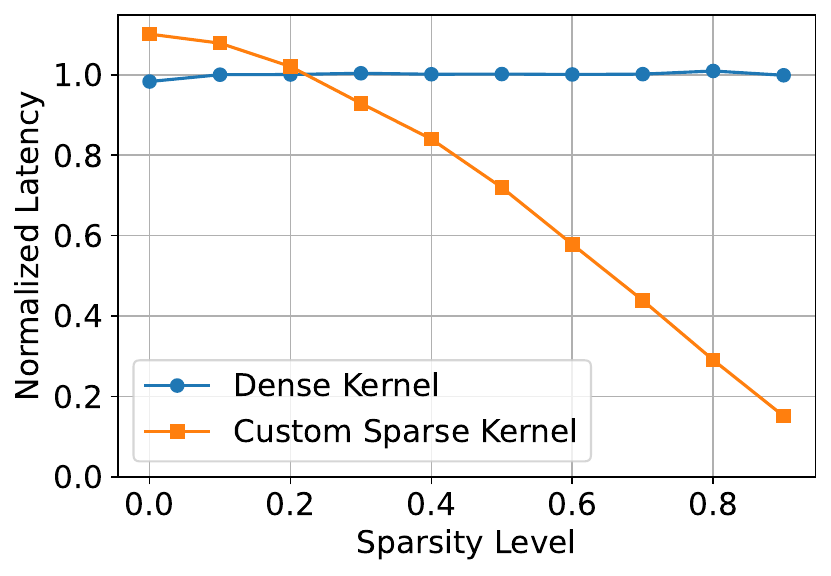}}
\subfigure[Up projection (\textit{vmmsp} kernel)]{\includegraphics[width=0.32\linewidth]{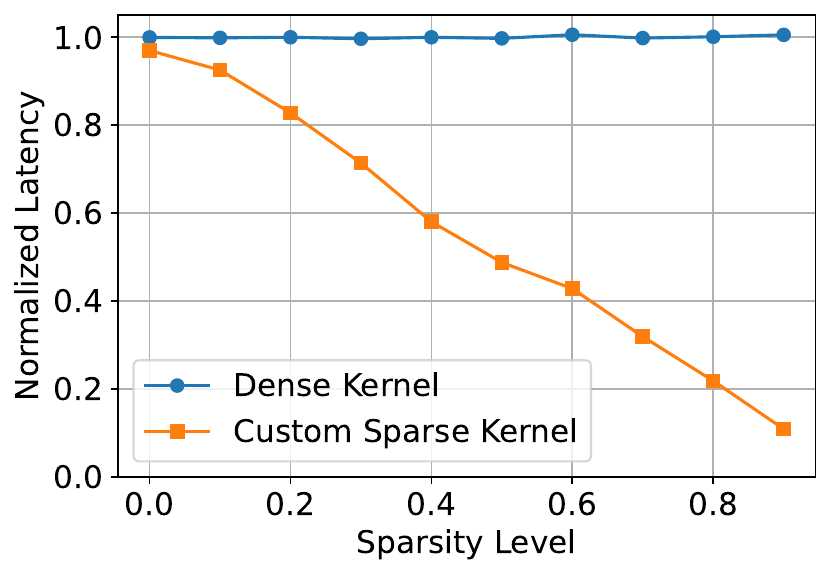}}
\end{tabular}
\caption{Comparison between custom sparse kernels and PyTorch dense kernel on latency of linear projections}
\label{fig:kernel_latency}
\end{figure*}

\begin{figure}[tbh]
    \centering
    \includegraphics[width=.9\linewidth]{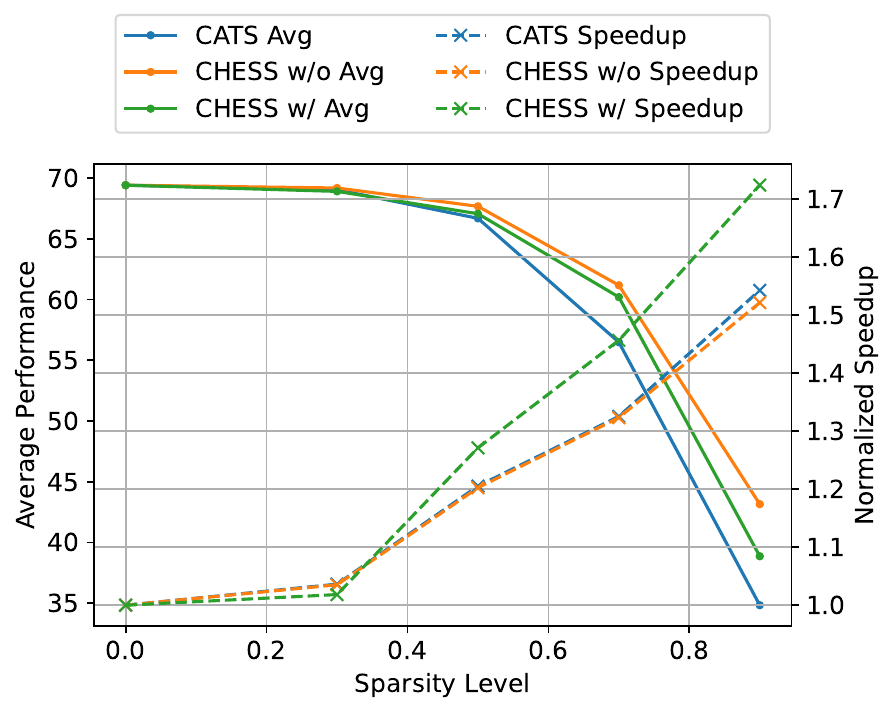}
    \caption{Downstream performance and end-to-end speedups of each method under different sparsity levels.}
    \label{fig:sparsity_vs_performance}
\end{figure}

\subsection{Main Results on Downstream Tasks}
\label{sec:downstream_performance}

Table~\ref{tab:performance_comparison} compares the accuracy of different models across eight downstream tasks and Figure~\ref{fig:end_to_end} evaluates the end-to-end inference speedups. 
Experimental results draw the following conclusions.

\textbf{Channel-wise thresholding can reduce accuracy degradation while achieving comparable sparsity.} 
Achieving a comparable sparsity, the proposed \name w/o exhibits a smaller average performance drop of 1.07 across five base models and eight downstream tasks, compared to the 1.70 degradation of CATS.
Specifically, \name w/o consistently outperforms CATS on ARC Easy, BoolQ, and HellaSwag, while showing modest gains on the remaining benchmarks.

\textbf{Selective sparsification of attention modules further improves sparsity while maintaining model accuracy.} 
Compared to \name w/o, the average performance of \name w/ degrades by 0.04 on Llama-2-7B and 0.61 on Llama-3-8B, respectively. 
Interestingly, for larger models such as Llama-2-13B, Llama-2-70B, and Mistral-7B, \name w/ demonstrates comparable or even slightly superior overall performances.
Specifically, \name w/ outperforms on OpenbookQA, but underperforms on ARC Easy, HellaSwag and BoolQ, while showing similar results on ARC Challenge, PIQA, SciQ, and Winogrande.
These results demonstrate that the additional selective sparsification on attention modules has minimal impact on performance.
In comparison to CATS, \name w/ consistently delivers superior average performance with fewer activated parameters.

\subsection{End-to-End Decoding Speedup}

\textbf{\name achieves end-to-end speedups of up to 1.27x compared to Transformers baselines.} 
When not employing attention sparsification, \name w/o achieves comparable speedups to CATS, which is 1.17x on Llama-2-7B and Llama-2-13B, 1.20x on Llama-3-8B, and 1.21x on Mistral-7B, respectively. 
This is because of the comparable parameters activated per decoding pass of these two methods.
When employing attention sparsification, the proposed \name w/ achieves the highest speedup of 1.25x on Llama-2-7B and Llama-2-13B, and 1.27x on Llama-3-8B and Mistral-7B, respectively.
Due to the limited capacity of main memory of edge devices, we did not perform the end-to-end speedup experiment for the Llama-2-70B model. 
However, based on the activated parameter count per inference pass, its speedup is estimated to be similar to that of Mistral-7B and Llama-3-8B.

\subsection{Ablation Study}

\label{sec:fs_vs_ss}

Table \ref{tab:ablation} presents the ablation study with different sparsification in attention modules.
While selective sparsification achieves a comparable reduction in overhead relative to full sparsification, it significantly outperforms full sparsification across all eight benchmarks.
Specifically, selective sparsification exhibits substantial improvements on the HellaSwag and Arc Challenge benchmarks, while demonstrating modest gains on the remaining benchmarks. 
These results underscore the advantages of selective sparsification.

\subsection{Kernel Efficiency}

As illustrated in Figure \ref{fig:kernel_latency}, this paper compares the latency against sparsity level between the proposed custom sparse kernel and the dense kernel in PyTorch~\citep{pytorch}.
At a sparsity level of 0, the \textit{vmmsp} kernel used for up projections demonstrates slightly lower latency compared to the PyTorch dense kernel. 
Conversely, the \textit{spvmm} kernel, utilized by attention projections and down projections, exhibits slightly higher latencies than the dense kernel. 
This increased latency is primarily due to the advanced loop tiling and reordering strategies, which cause slight performance degradation at low sparsity levels.

As the sparsity level increases, the latency of the dense kernel remains relatively constant, whereas the latency of our custom sparse kernels decreases proportionally. Notably, at a sparsity level of 0.5, our custom sparse kernels achieve latency reductions of 30\%, 28\%, and 51\% for attention projection, FFN up projection, and FFN down projection, respectively. These findings highlight the efficiency of our custom kernels.

\subsection{Impact on Different Sparsity Levels}

Figure \ref{fig:sparsity_vs_performance} shows the model performance on downstream tasks and end-to-end decoding speedups at different sparsity levels. 
We selected Llama-3-8B as the base model since it incorporates the contemporary GQA module.

Experimental results indicate that at lower sparsity levels (0.3 and 0.5), both CATS and \name maintain performance comparable to the base model, with \name exhibiting superior performance. 
At higher sparsity levels (0.7 and 0.9), these models experience noticeable performance degradation, and \name models, particularly \name w/o models, consistently outperform CATS. 
Specifically, at a sparsity level of 0.7, the CATS, \name w/o, and \name w/ models achieve average performances of 56.49, 61.18, and 60.21, respectively. 
At a sparsity level of 0.9, the corresponding performances are 34.83, 43.15, and 38.86, respectively.

Regarding end-to-end speedup, \name w/ exhibits the highest speedup at all sparsity levels above 0.3, attributed to the selective sparsification of attention modules. 
Specifically, \name w/ achieves speedups of 1.46x and 1.72x at sparsity levels of 0.7 and 0.9, respectively, compared to 1.33x and 1.52x for CATS. 
However, at a sparsity level of 0.3, the \name w/ exhibits slightly reduced speedup, mainly due to the suboptimal efficiency of the \textit{spvmm} kernel at lower sparsity levels.

\subsection{Extended Comparisons with State-of-the-Art Training-Free Pruning Methods}

To further demonstrate the effectiveness of our proposed \name method, we extend our comparisons to include other state-of-the-art training-free pruning approaches, such as Relufication \citep{relu_strikes_back} and Wanda \citep{wanda}.
Notably, although Relufication achieves competitive performance when fine-tuned, it struggles with performance degradation in training-free scenarios. 
Wanda, on the other hand, focuses on weight pruning, which belongs to a different branch of work. 
Weight pruning typically results in unstructured sparsity or semi-structured sparsity, which is only supported by high-end NVIDIA GPUs with Ampere or Hopper architectures. 
In contrast, our proposed CHESS does not rely on specialized GPU architecture, making it more suitable for deploying on edge devices.

As presented in Table \ref{tab:more_comparison}, the proposed CHESS method achieves superior performance in most benchmarks while activating comparable or fewer parameters compared to both Relufication and Wanda.
Specifically, CHESS with a sparsity level of 0.7 outperforms other methods on several benchmarks including Arc Challenge, Arc Easy, HellaSwag, OpenbookQA, PIQA and Winogrande. 
Despite using only 54.92\% of the model’s parameters per decoding pass, CHESS delivers an average performance (60.21) that surpasses Wanda (56.41) and Relufication (28.94).
These results emphasize the advantage of CHESS over existing methods.

%% file: sections/6-rel.tex
\section{Related Work}

Various methods have been proposed to address the challenges associated with deploying LLMs locally. Weight quantization \citep{smoothquant, gptq, awq} aims to represent LLM weights using lower bit-widths, thereby reducing memory usage and access overhead. 
Activation quantization focuses on minimizing the memory footprint of activation tensors and KV cache ~\citep{atom, kivi, kvquant}. These methods can be applied along with our proposed \name method.

Weight pruning \citep{sparsegpt, wanda} involves setting a portion of the LLM weights to zero to reduce computational overhead and memory requirement. However, this approach faces several challenges including noticeable degradation in performance and limited hardware support when applied on personal devices.

Non-autoregressive decoding approaches, such as speculative decoding \citep{speculative_decoding, distillspec} or Medusa \citep{medusa}, seek to convert autoregressive decoding process of LLMs into parallel decoding to mitigate memory access overhead. However, these methods impose increased computational demands, which presents significant challenges for deployment on personal devices with limited processing capabilities.

%% file: sections/5-conclusions.tex
\section{Conclusion}

This paper reformulates the activation sparsification problem and 
introduces the \name, a general activation sparsification via channel-wise thresholding and selective sparsification.
Experiments show that the proposed \name can achieve a lower performance degradation and accelerate the LLM inference with sparse activations.

%% file: sections/7-lim.tex
\section*{Limitations}

The limitations of this work can be summarized in two main aspects. 
First, while \name achieves lower accuracy degradation compared to existing methods with fewer activated parameters, it still experiences a noticeable accuracy loss at higher sparsity levels. 
Future research could explore fine-tuning techniques to mitigate this decline in performance.
Second, \name is optimized for inference with a batch size of one, which is suitable for edge deployment scenarios typically involving a single user. 
However, under larger batch sizes, the structured sparsity of activation tensors deteriorates into unstructured sparsity, limiting potential speedups and reducing effectiveness in data center deployments, where larger batch sizes are common.

%% file: sections/8-ack.tex
\section*{Acknowledgements}
We thank all the reviewers for their insightful comments. This work is supported by the National Natural Science Foundation of China (No. 62472330, U20A20177, 62272348, U22B2022), the National Key Research and Development Program of China (No. 2022YFB3104502), the State Key Laboratory of Computer Architecture (ICT, CAS) under Grant No. CARCH A202112, and Wuhan Science and Technology Joint Project for Building a Strong Transportation Country (No.2023-2-7).